\documentclass{article}
\usepackage{xcolor,colortbl}
\usepackage{arxiv}
\usepackage{amsmath}
\usepackage{tikz}
\usepackage{chemfig}
\usepackage{marvosym}
\usetikzlibrary{arrows.meta}

\usepackage[utf8]{inputenc} 
\usepackage[T1]{fontenc}    
\usepackage{hyperref}       
\usepackage{url}            
\usepackage{booktabs}       
\usepackage{multirow}
\usepackage{amsfonts}       
\usepackage{nicefrac}       
\usepackage{microtype}      
\usepackage{lipsum}
\usepackage{graphicx}
\graphicspath{ {./images/} }
\usepackage{graphicx}

\usepackage{scrextend}
\usepackage{pgfplots}

\setchemfig{chemfig style={line width=0.5pt}}
\setchemfig{atom style={scale=0.5}}

\newtheorem{definition}{Definition}

\usetikzlibrary{positioning}
\usepackage{tikz-3dplot,calc}
\usepackage[labelfont=bf,justification=centering]{caption}

\definecolor{atomictangerine}{rgb}{1.0, 0.6, 0.4}
\definecolor{blue-green}{rgb}{0.0, 0.87, 0.87}
\definecolor{aquamarine}{rgb}{0.5, 1.0, 0.83}
\definecolor{babypink}{rgb}{0.96, 0.76, 0.76}
\definecolor{fluorescentyellow}{rgb}{0.8, 1.0, 0.0}
\definecolor{harlequin}{rgb}{0.25, 1.0, 0.0}
\definecolor{brightgreen}{rgb}{0.4, 1.0, 0.0}
\definecolor{coolgray}{rgb}{0.55, 0.57, 0.67}

\DeclareFontFamily{OT1}{pzc}{}
\DeclareFontShape{OT1}{pzc}{m}{it}{<-> s * [1.10] pzcmi7t}{}
\DeclareMathAlphabet{\mathpzc}{OT1}{pzc}{m}{it}

\usepackage{float}

\title{FunQG: Molecular Representation Learning Via Quotient Graphs}

\author{ 
Hossein Hajiabolhassan\footnotemark[1] $^{\ ,}$\footnotemark[2] $^{\ ,}$\footnotemark[4] , 
\ Zahra Taheri\footnotemark[2] $^{\ ,}$\footnotemark[4] , 
\ Ali Hojatnia\footnotemark[3] $^{\ ,}$\footnotemark[4] , 
\ Yavar Taheri Yeganeh\footnotemark[3] $^{\ ,}$\footnotemark[4] \\ 
}

\begin{document}

\maketitle
\newcounter{mycounter}
\deffootnote[0em]{0em}{0em}{%
  \makebox[0em][l]{\textsuperscript{\thefootnotemark}}%
}
\renewcommand{\thefootnote}{\fnsymbol{mycounter}}
\footnotetext[1]{\makeatletter\setcounter{footnote}{0}
\renewcommand{\thefootnote}{\fnsymbol{footnote}}{\footnotemark[1]}\makeatother \ Corresponding author: Department of Mathematics and Information Technology, Chair of Information Technology, \newline \hspace*{0.8em}Montanuniversit\"{a}t Leoben, Franz-Josef-Strasse 18, A-8700 Leoben, Austria. E-mail: \href{email:hossein.hajiabolhassan@unileoben.ac.at}{hossein.hajiabolhassan@unileoben.ac.at}}

\footnotetext[4]{\makeatletter\setcounter{footnote}{0}
\renewcommand{\thefootnote}{\fnsymbol{footnote}}{\footnotemark[4]}\makeatother \ Machine Learning and Graph Mining Lab, Department of Applied Mathematics, Faculty of Mathematical Sciences, Shahid \newline \hspace*{0.7em} Beheshti University, 19839-69411, Tehran, Iran.}

\footnotetext[2]{\makeatletter\setcounter{footnote}{0}
\renewcommand{\thefootnote}{\fnsymbol{footnote}}{{\footnotemark[2]}$^{,}${\footnotemark[3]}}\makeatother \ These two authors have contributed equally to this work and their names are listed alphabetically.}
\setcounter{footnote}{0}\renewcommand{\thefootnote}{\arabic{footnote}}

\begin{abstract}
Learning expressive molecular representations is crucial to facilitate the accurate prediction of molecular properties. Despite the significant advancement of graph neural networks (GNNs) in molecular representation learning, they generally face limitations such as neighbors-explosion, under-reaching, over-smoothing, and over-squashing. Also, GNNs usually have high computational costs because of the large-scale number of parameters. Typically, such limitations emerge or increase when facing relatively large-size graphs or using a deeper GNN model architecture. An idea to overcome these problems is to simplify a molecular graph into a small, rich, and informative one, which is more efficient and less challenging to train GNNs. To this end, we propose a novel molecular graph coarsening framework named \textbf{FunQG} utilizing \textbf{Fun}ctional groups, as influential building blocks of a molecule to determine its properties, based on a graph-theoretic concept called \textbf{Q}uotient \textbf{G}raph. By experiments, we show that the resulting informative graphs are much smaller than the molecular graphs and thus are good candidates for training GNNs. We apply the FunQG on popular molecular property prediction benchmarks and then compare the performance of some popular baseline GNNs on the obtained datasets with the performance of several state-of-the-art baselines on the original datasets. By experiments, this method significantly outperforms previous baselines on various datasets, besides its dramatic reduction in the number of parameters and low computational costs. Therefore, the FunQG can be used as a simple, cost-effective, and robust method for solving the molecular representation learning problem.~
\end{abstract}

\keywords{Deep learning; Molecular representation; Molecular property; Graph neural network; Quotient graph; Functional group}


\begin{figure*}[ht!]
\begin{center}
\hspace*{-1em}
\includegraphics[scale=0.7]{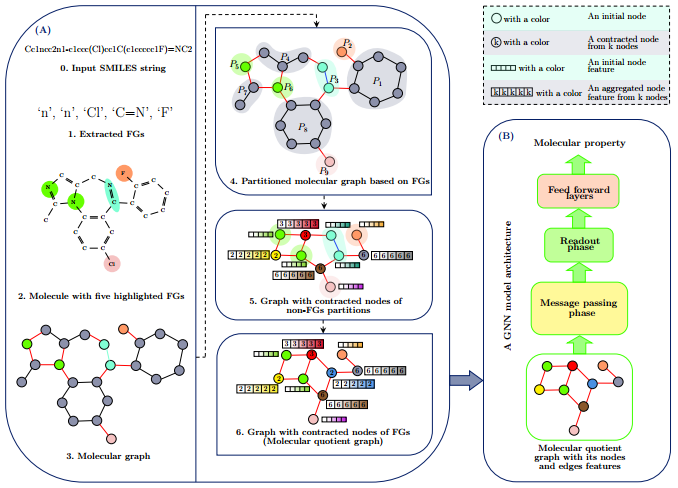}
\vspace*{0.3em}
\end{center}
\caption{\textbf{The overview of FunQG framework.} The left figure (A) illustrates the application of the FunQG framework to a molecule to find its corresponding coarsened graph, named molecular quotient graph. The right figure (B) shows the application of a GNN architecture to the graph obtained from the FunQG to predict the property of the molecule. In the molecular graph, a specific color corresponds to each FG (its edges and nodes). Also, edges that have exactly one common node with an FG are drawn in red. The remained edges are drawn in black. }\label{fig-qg-2}
\end{figure*}

\section{Introduction}\label{sec:intro}
Drug discovery and development are often fraught with problems such as high costs and time-consuming processes, e.g., the production of a drug by traditional methods costs an average of $2.6$ billion USD and can take more than $12$ years~\cite{Chan}. Innovative approaches are essential to reduce the costs and accelerate the development. Computational methods can improve the drug discovery process due to their rapid and accurate prediction. Artificial intelligence is among solutions to drug development, especially in predicting molecular properties~\cite{Walters}, drug-drug interaction~\cite{Ryu}, and drug-target interaction~\cite{Abbasi}. The performance of its methods relies heavily on finding the expressive representations of molecular structures~\cite{Yang}. In the early works, molecular representations are based on fixed expert-designed molecular descriptors or fingerprints, such as Dragon descriptors or Morgan (ECFP) fingerprints~\cite{Mauri, Rogers}. On the other hand, domain-specific expertise is used to select or engineer molecular descriptors~\cite{Cao, Durant, Moriwaki}. 

The highly flexible architecture of neural networks enables them to adapt to various problems and achieve remarkable gains by extracting complex relationships. Instead of relying on features compiled by an expert, neural models can learn expressive representations directly from raw data~\cite{Coley, Duvenaud, Kearnes}.  

\textit{Graph neural networks} (\textit{GNNs}), which have been developed as a powerful candidate for graph-structured data modeling~\cite{Gori, Micheli, Scarselli}, are among the hottest topics in recent years~\cite{Duvenaud, Hamilton, Xu}. A molecule can be represented naturally as a graph in which atoms (nodes) connect by chemical bonds (edges). It has been shown in numerous studies that graph-based molecular property prediction can be more accurate than traditional descriptor-based methods~\cite{Coley, Duvenaud, Gilmer, Kearnes, Xiong}. On the other hand, some other studies have shown that utilizing only computed descriptors achieved better results than graph-based models~\cite{Jiang}. Therefore, each has its own benefits~\cite{Yang}. Accordingly, some works that have followed a hybrid approach, incorporating easily computed descriptors with a graph-based model, have shown promising performance in drug discovery over non-neural models~\cite{Stokes, Yang}. 

The key idea of GNNs is the use of a neural message passing process, in which message vectors of the nodes\textquotesingle~neighbors (or edges\textquotesingle~neighbors) are aggregated through neural networks to update node (and/or edge) representations~\cite{Gilmer, Kipf, Veli, Yang}. Therefore, a GNN layer combines the independent information of each node (and/or edge) with the structural information obtained from its neighbors. However, most learning problems require multiple GNN layers to exchange information between nodes (and/or edges) that are not directly connected. 

Although existing GNN architectures have made good progress in various fields, they also face limitations that need further investigation. Some of such limitations are common problems observed in neural networks, such as overfitting. Others are structural limitations specific to graphs that typically emerge when the scale of the graph increases or the GNN model architecture goes deeper. The process of recursive aggregation of neighborhoods in GNNs causes the explosion of the number of neighbors and thus leads to exponentially-growing computation complexity due to the use of multiple GNN layers. This problem which is described as \textit{neighbors-explosion} results in serious memory and computation bottlenecks, especially in working with large-size graphs~\cite{Bai}. On the other hand, if the number of GNN layers is less than the diameter of the graph, then nodes that are farther away than the number of GNN layers are unable to receive information from each other~\cite{Barcel}. Alon~et~al.~\cite{Alon} refer to this limitation as \textit{under-reaching}. Another limitation is due to the \textit{over-smoothing} phenomenon, in which the node representations become almost indistinguishable by increasing the number of GNN layers~\cite{Li, Oono, Wu}. However, over-smoothing is more common in short-range problems, such as prediction tasks on social networks~\cite{Leskovec}, citation networks~\cite{Sen}, and product recommendations~\cite{Shchur}. The efficiency in such cases usually depends only on short-range information from each node\textquotesingle s local neighborhood and does not improve by adding distant information. In contrast, \textit{over-squashing} of information from exponentially many neighbors into fixed-size vectors is a performance bottleneck in tasks that require long-range interaction and so more number of GNN layers~\cite{Alon}, such as molecular property prediction tasks, which may depend on the combination of atoms that reside in the molecule\textquotesingle s opposite sides~\cite{Matlock}. To avoid under-reaching, the number of GNN layers should be as many as the range of interactions. But as the depth of GNNs increases, the neighbors-explosion, over-smoothing, and over-squashing become more likely. Using GNNs whose number of layers was more than the diameter of the graphs, experimentally on chemical property prediction tasks, Alon~et~al.~\cite{Alon} show that over-squashing is the cause of the poor performance of such neural networks, not under-reaching.

Different approaches have been investigated to improve the performances of GNNs in molecular property prediction tasks. Common strategies include proposing a new GNN variant, self-supervised pre-training, and pre-processing raw input data. Each of such approaches has its advantages and disadvantages. For example, despite acceptable performances of self-supervised pre-training methods~\cite{Fang, Li2, Ma, Rong}, these methods have high computational costs and require rich hardware resources. The main focus of this paper is proposing a specific type of pre-processing method for molecular data that not only focuses on solving the above issues of GNNs in property prediction tasks but also reduces computational costs. In order to achieve this goal, an idea is to simplify the graphs in such a way that by using fewer GNN layers to the resulting small graphs, not only memory and computational costs are reduced but also significant performances are achieved. Typical approaches for simplifying graphs can be divided into two major categories. In the first category, graph edge \textit{sparsification methods} approximate a graph to a sparse graph by reducing the number of edges~\cite{Peleg,Spielman1,Spielman2}. In the second category, graph \textit{coarsening methods} reduce the number of nodes to a subset of the original node set~\cite{Bravo, Cai, Loukas1, Loukas2}. In this paper, we propose a novel molecular graph coarsening method, which reduces the number of nodes by contracting nodes in certain subgraphs. Such subgraphs are obtained from a partition of the molecular graph. Our approach is based on a concept of graph theory called \textit{quotient graph} (see Definition \ref{def-qg}), and the resulting coarsened graph is named \textit{molecular quotient graph}. 

Although there are various methods for graph coarsening, most are not data-driven or even domain-specific~\cite{Cai}. Also, the purpose of most existing graph coarsening algorithms is to maintain some graph properties such as principal eigenvalues~\cite{Loukas1}, which may be suboptimal. The main focus of this paper is to propose a coarsening method specific to molecular graphs.  Specifically, we aim to miniaturize the molecular graph into a rich and informative one such that GNNs with fewer layers trained on the resulting small graphs have great efficiency and even better performance than GNNs with more layers trained on the molecular graphs. 

Since the process of constructing quotient graphs is prone to information loss, selecting appropriate graph partitions is crucial to maintain information in the resulting quotient graphs. Different methods can be considered to partition the node set of a molecular graph. But based on our experiments, we have decided to use the domain knowledge of chemistry for this case. On the other hand, unlike a molecule, a graph is a 2-dimensional structure with no spatial relationship between its elements. Yet, some 3-dimensional information of a molecule and the information resulting from its 3-dimensional structure, such as stereochemistry, can be encoded in its feature vectors of the nodes and edges~\cite{David}. In this paper, to produce an expressive molecular representation and avoid data loss, information in feature vectors of the nodes and edges of a molecular graph are maintained in the feature vectors of the nodes and edges of its molecular quotient graph by an appropriate local aggregation method.

It is known that a molecule may consist of many atomic groups, while certain atomic groups, called \textit{functional groups} (\textit{FGs}), determine the properties and reactivity of the parent molecule~\cite{Ertl, Muller}. Despite the importance of FGs in predicting molecular properties, most existing GNN models completely ignore them. Various substructure features, often used in cheminformatics, are usually generated based on fragments or local properties of atoms and bonds. In general, the fragments are strongly overlapping and are generated for all parts of a molecule without considering their potential chemical role~\cite{Ertl}. Therefore, these features do not describe FGs. In this paper, we propose a framework for the molecular graph coarsening, named \textit{FunQG}, which firstly partitions the molecular graph\textquotesingle s nodes based on FGs and then contracts the nodes of each partition in such a way that information is preserved in the new nodes and edges. This pre-processing method is independent of GNN architectures. By our experiments, the resulting informative molecular quotient graphs are much smaller than the molecular graphs (see Supplementary Section 4). Because many of the problems of GNNs are rooted in working with relatively large-size graphs, the resulting small molecular quotient graphs are good candidates to be used more efficiently and with fewer challenges for training GNNs. As far as we know, this molecular graph coarsening framework is the first approach that uses information about FGs to find more expressive molecular representations for predicting molecular properties by deep learning. To demonstrate the effectiveness of the proposed graph coarsening framework, we combine FunQG with two different famous GNN model architectures,~MPNN~\cite{Gilmer} and DMPNN~\cite{Yang}, and call the resulting combinations \textit{FunQG-MPNN} and \textit{FunQG-DMPNN}, respectively. In FunQG-MPNN (FunQG-DMPNN), we firstly apply the FunQG on a molecular property prediction dataset, and then we compare the performance of MPNN (DMPNN) on the obtained dataset with the performances of several baselines on the original dataset. Based on our experiments on several molecular property prediction benchmarks, the FunQG-MPNN (FunQG-DMPNN) inclusively outperforms the vanilla MPNN (vanilla DMPNN). Also, the experimental results show that the FunQG-DMPNN significantly outperforms previous baselines on various drug discovery tasks.

\section{Materials and Methods}
\label{sec:method}

\subsection{Preliminaries}
Throughout this paper, suppose that graphs are finite, undirected (or equivalently, directed in which the reverse of every edge is also an edge), and simple (loopless with no parallel edges). Also, if $G$ is a graph, then the node set of $G$ is denoted by $V(G)$, and its edge set is denoted by $E(G)$. If two nodes $v$ and $w$ are adjacent in $G$, then it is denoted by $vw\in E(G)$ (or equivalently by $(v,w)\in E(G)$ as a directed edge from $v$ to $w$). Furthermore, the set of neighbors of a node $v$ is denoted by $N(v)$.

A molecule $\mathcal{M}$ can be represented as a graph $G_\mathcal{M}$ in which every node $v$ is corresponding to an atom $\mathpzc{v}$ in $\mathcal{M}$, and for all $v,w\in V(G_\mathcal{M})$, $vw\in E(G_\mathcal{M})$ if and only if there exists a chemical bond between atoms $\mathpzc{v}$ and $\mathpzc{w}$ in $\mathcal{M}$. The initial features of a node $v$ is denoted by $\mathbf{x}_v$, and the initial features of an edge $(v,w)$ is denoted by $\mathbf{e}_{vw}$. For initial feature extraction, we follow the same protocol as Wu~et~al.~\cite{Wu2} and Yang~et~al.~\cite{Yang}. The detailed feature extraction process and initial atom and bond features can be found in the Supplementary Section 3. 

\subsubsection{Message Passing Neural Networks} 
GNNs, utilizing the node and edge features along with the graph structure as inputs, use a neural message passing process to learn the representation vector of the nodes. More precisely, in the message passing phase, the independent information of each node $v$ with the structural information of its neighbors are combined iteratively to obtain a representation vector $h_v$ of the node. Then in the readout phase, the representation vector $h_G$ of the graph is obtained by pooling over the final representation vectors of its nodes to make predictions about the properties of interest. 

\textbf{MPNN.} An \textit{MPNN} operates on an undirected graph $G$. On each step $t\in \{0,1,\ldots,T-1\}$ of the message passing phase, for all nodes $v$, the hidden states $h_v^t$ and the messages $m_v^t$ are updated as follows
\begin{align*}
m_v^{t+1} &= \sum_{w\in N(v)} M_t(h_v^t,h_w^t,\mathbf{e}_{vw}), \\
h_v^{t+1} &= U_t(h_v^t,m_v^{t+1}), 
\end{align*}
where $M_t$, $U_t$, and $h_v^0$ are respectively a message function, a vertex update function, and a function of the initial node features $\mathbf{x}_v$. Afterwards, in the readout phase, a readout function $R$ is used to predict the property $\hat{y} = R(\{h_v^T \vert v\in V(G)\})$. 

To have a fair comparison, our implementation of MPNN is the same as that of Yang~et al.~\cite{Yang}, in which learnable parameters are the weight matrices $W_i\in \mathbb{R}^{h\times n_i}$, $W_m\in \mathbb{R}^{h\times (h+e_i)}$, and $W_o\in \mathbb{R}^{h\times (h+n_i)}$, where $n_i$ is the initial node features size, $e_i$ is the initial edge features size, and $h$ is called the \textit{hidden size}. Also, $h_v^0=W_i\mathbf{x}_v$, $M_t(h_v^t,h_w^t,\mathbf{e}_{vw})=\text{cat}(h_w^t, \mathbf{e}_{wv})$, and $U_t(h_v^t,m_v^{t+1})=\tau(h_v^0+W_m m_v^{t+1})$, where $\tau$ is the \textit{ReLU} activation function. Then, node representations $h_v$ are obtained as follows
\begin{align}
m_v &= \sum_{w\in N(v)} h_{w}^T,\label{eq-5}\\
h_v &= \tau(W_o \text{ cat}(\mathbf{x}_v,m_v)). \label{eq-6}
\end{align}
Furthermore, the average or sum pooling function is utilized in the readout phase to obtain the graphs representations $h_G$. Then a feed-forward neural network $f$ is used to generate property predictions $\hat{y} = f(h_G)$. 

\begin{figure}[t]
	\begin{center}
		\usetikzlibrary{shapes.geometric}
		\hspace*{0em}{
\tikzset{every picture/.style={line width=0.75pt}} 

\begin{tikzpicture}[x=0.75pt,y=0.75pt,yscale=-0.65,xscale=0.65]

\draw  [fill=blue-green  ,fill opacity=1 ] (151,44.5) .. controls (151,36.49) and (157.49,30) .. (165.5,30) .. controls (173.51,30) and (180,36.49) .. (180,44.5) .. controls (180,52.51) and (173.51,59) .. (165.5,59) .. controls (157.49,59) and (151,52.51) .. (151,44.5) -- cycle ;
\draw  [fill=blue-green  ,fill opacity=1 ] (231,96.5) .. controls (231,88.49) and (237.49,82) .. (245.5,82) .. controls (253.51,82) and (260,88.49) .. (260,96.5) .. controls (260,104.51) and (253.51,111) .. (245.5,111) .. controls (237.49,111) and (231,104.51) .. (231,96.5) -- cycle ;
\draw  [fill=blue-green  ,fill opacity=1 ] (230,165.5) .. controls (230,157.49) and (236.49,151) .. (244.5,151) .. controls (252.51,151) and (259,157.49) .. (259,165.5) .. controls (259,173.51) and (252.51,180) .. (244.5,180) .. controls (236.49,180) and (230,173.51) .. (230,165.5) -- cycle ;
\draw  [fill=atomictangerine  ,fill opacity=1 ] (151,214.5) .. controls (151,206.49) and (157.49,200) .. (165.5,200) .. controls (173.51,200) and (180,206.49) .. (180,214.5) .. controls (180,222.51) and (173.51,229) .. (165.5,229) .. controls (157.49,229) and (151,222.51) .. (151,214.5) -- cycle ;
\draw  [fill=atomictangerine  ,fill opacity=1 ] (71,164.5) .. controls (71,156.49) and (77.49,150) .. (85.5,150) .. controls (93.51,150) and (100,156.49) .. (100,164.5) .. controls (100,172.51) and (93.51,179) .. (85.5,179) .. controls (77.49,179) and (71,172.51) .. (71,164.5) -- cycle ;
\draw  [fill=harlequin  ,fill opacity=1 ] (72,94.5) .. controls (72,86.49) and (78.49,80) .. (86.5,80) .. controls (94.51,80) and (101,86.49) .. (101,94.5) .. controls (101,102.51) and (94.51,109) .. (86.5,109) .. controls (78.49,109) and (72,102.51) .. (72,94.5) -- cycle ;
\draw [color=black  ,draw opacity=1 ][line width=1.1]    (178,52.33) -- (233,88) ;
\draw [color=black  ,draw opacity=1 ][line width=1.1]    (96,173.8) -- (152,208) ;
\draw [color=black  ,draw opacity=1 ][line width=1.1]    (245.5,111) -- (244.5,151) ;
\draw [color=blue  ,draw opacity=1 ][line width=1.1]    (234,175.4) -- (178,207.4) ;
\draw  [color={rgb, 255:red, 0; green, 0; blue, 0 }  ,draw opacity=0 ][fill=blue-green  ,fill opacity=0.3 ] (144.73,29.93) .. controls (158.38,15.64) and (197.93,31.97) .. (223.27,45.72) .. controls (238.07,53.76) and (248.01,60.91) .. (245.1,60.57) .. controls (237.19,59.64) and (261.1,59.57) .. (271.1,101.57) .. controls (281.1,143.57) and (271.1,216.57) .. (241.1,196.57) .. controls (211.1,176.57) and (123.1,52.57) .. (144.73,29.93) -- cycle ;
\draw [color=black  ,draw opacity=1 ][line width=1.1]    (173.1,57.57) -- (234.1,154.57) ;
\draw [color=blue  ,draw opacity=1 ][line width=1.1]    (165.5,59) -- (165.5,200) ;
\draw [color=green  ,draw opacity=1 ][line width=1.1]    (95.1,107) -- (158,202) ;
\draw [color=blue  ,draw opacity=1 ][line width=1.1]    (100,164.5) -- (230,165.5) ;
\draw  [color={rgb, 255:red, 0; green, 0; blue, 0 }  ,draw opacity=0 ][fill=atomictangerine  ,fill opacity=0.3 ] (122.55,157.93) .. controls (161,184) and (205,213) .. (185,232) .. controls (165,251) and (110,214) .. (104,211) .. controls (98,208) and (56.55,173.93) .. (62,155) .. controls (67.45,136.07) and (84.1,131.85) .. (122.55,157.93) -- cycle ;
\draw  [color={rgb, 255:red, 0; green, 0; blue, 0 }  ,draw opacity=0 ][fill=harlequin  ,fill opacity=0.3 ] (45.08,87.5) .. controls (45.08,68.49) and (60.49,53.08) .. (79.5,53.08) .. controls (98.51,53.08) and (113.92,68.49) .. (113.92,87.5) .. controls (113.92,106.51) and (98.51,121.92) .. (79.5,121.92) .. controls (60.49,121.92) and (45.08,106.51) .. (45.08,87.5) -- cycle ;
\draw  [fill=blue-green  ,fill opacity=1 ] (483,74.5) .. controls (483,63.18) and (492.18,54) .. (503.5,54) .. controls (514.82,54) and (524,63.18) .. (524,74.5) .. controls (524,85.82) and (514.82,95) .. (503.5,95) .. controls (492.18,95) and (483,85.82) .. (483,74.5) -- cycle ;
\draw  [fill=atomictangerine  ,fill opacity=1 ] (403,193.5) .. controls (403,182.18) and (412.18,173) .. (423.5,173) .. controls (434.82,173) and (444,182.18) .. (444,193.5) .. controls (444,204.82) and (434.82,214) .. (423.5,214) .. controls (412.18,214) and (403,204.82) .. (403,193.5) -- cycle ;
\draw  [fill=harlequin  ,fill opacity=1 ] (375,73.5) .. controls (375,62.18) and (384.18,53) .. (395.5,53) .. controls (406.82,53) and (416,62.18) .. (416,73.5) .. controls (416,84.82) and (406.82,94) .. (395.5,94) .. controls (384.18,94) and (375,84.82) .. (375,73.5) -- cycle ;
\draw [color=green  ,draw opacity=1 ][line width=1.1]    (395.5,95) -- (417,173) ;
\draw [color=blue  ,draw opacity=1 ][line width=1.1]    (492.5,91) -- (436,177.71) ;
\draw  [color={rgb, 255:red, 74; green, 74; blue, 74 }  ,draw opacity=1 ][fill={rgb, 255:red, 224; green, 151; blue, 236 }  ,fill opacity=0.8 ][line width=0.75]  (290.07,117.75) -- (344.63,117.75) -- (344.63,103) -- (381,132.5) -- (344.63,162) -- (344.63,147.25) -- (290.07,147.25) -- cycle ;

\draw (78,157) node [anchor=north west][inner sep=0.75pt]   [align=left] {1};
\draw (79,86) node [anchor=north west][inner sep=0.75pt]   [align=left] {2};
\draw (159,36) node [anchor=north west][inner sep=0.75pt]   [align=left] {3};
\draw (238,88) node [anchor=north west][inner sep=0.75pt]   [align=left] {4};
\draw (237,156) node [anchor=north west][inner sep=0.75pt]   [align=left] {5};
\draw (158,206) node [anchor=north west][inner sep=0.75pt]   [align=left] {6};
\draw (491,66) node [anchor=north west][inner sep=0.75pt]   [align=left] {$\displaystyle V_{1}$};
\draw (411,185) node [anchor=north west][inner sep=0.75pt]   [align=left] {$\displaystyle V_{2}$};
\draw (383,65) node [anchor=north west][inner sep=0.75pt]   [align=left] {$\displaystyle V_{3}$};
\draw (159,240) node [anchor=north west][inner sep=0.75pt]   [align=left] {$\displaystyle G$};
\draw (409,238) node [anchor=north west][inner sep=0.75pt]   [align=left] {$\displaystyle G/\mathcal{P}$};
\draw (292,121) node [anchor=north west][inner sep=0.75pt]   [align=left] {Quotient};

\draw (55,60) node [anchor=north west][inner sep=0.75pt]   [align=left] {$\displaystyle V_{3}$};
\draw (100,190) node [anchor=north west][inner sep=0.75pt]   [align=left] {$\displaystyle V_{2}$};
\draw (200,50) node [anchor=north west][inner sep=0.75pt]   [align=left] {$\displaystyle V_{1}$};
\end{tikzpicture}
}
\caption{$G/\mathcal{P}$ is the quotient graph of $G$ under the equivalence relation specified by the partition set $\mathcal{P}=\{V_1, V_2, V_3\}$ of $V(G)$, where $V(G)=\{1,2,\ldots,6\}$, $V_1=\{3,4,5\}$, $V_2=\{1,6\}$, and $V_3=\{2\}$. By definition \ref{def-qg}, all nodes in each partition $V_i$ of $V(G)$ are contracted to a unique node $V_i$ in $G/\mathcal{P}$. Also, all blue (green) edges in $G$ are contracted to one blue (green) edge in $G/\mathcal{P}$.}\label{fig-qg}
\end{center}
\end{figure}
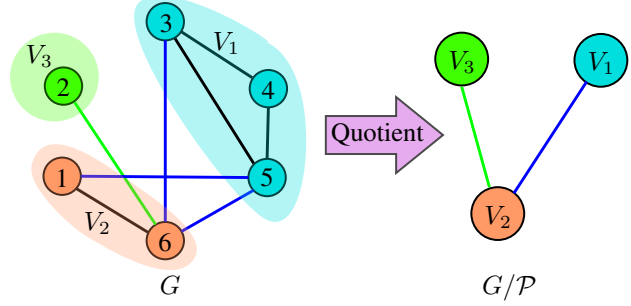


\textbf{Directed MPNN.} Unlike MPNN, in \textit{Directed MPNN} (\textit{DMPNN})~\cite{Dai, Yang} messages are propagated via directed edges in a way that prevents \textit{totters}~\cite{Mahe}, i. e., messages passed along a path of the form $v_1v_2\ldots v_n$ where $v_i = v_{i+2}$ for some $i$, because such paths are likely to introduce noise to the graph representation. This is achieved by updating messages $m_{vw}$ of the directed edge $(v,w)$ using all hidden states of the incoming edges $h_{kv}$, where $k\in N(v)\setminus w$.

The learnable parameters of the DMPNN~\cite{Yang} are the weight matrices $W_i\in \mathbb{R}^{h\times (n_i+e_i)}$, $W_m\in \mathbb{R}^{h\times h}$, and $W_o\in \mathbb{R}^{h\times (h+n_i)}$. In the message passing phase, the hidden states of an edge $(v,w)$ is initialized as
$$h_{vw}^0 = \tau(W_i \text{ cat}(\mathbf{x}_v,\mathbf{e}_{vw})),$$
where $\text{cat}(\mathbf{x}_v,\mathbf{e}_{vw})\in \mathbb{R}^{(n_i+e_i)}$ is the concatenation of the initial node features $\mathbf{x}_v$ and the initial edge features $\mathbf{e}_{vw}$, and $\tau$ is the \textit{ReLU} activation function. 
Then on each step $t\in \{0,1,\ldots,T-1\}$ of the message passing phase, for all directed edges $(v,w)$, the hidden states $h_{vw}^t$ and the messages $m_{vw}^t$ are updated as follows
\begin{align}
m_{vw}^{t+1} &= \sum_{k\in N(v)\setminus w} h_{kv}^t, \label{eq-1} \\
h_{vw}^{t+1} &= \tau(h_{vw}^0+W_m m_{vw}^{t+1}). \label{eq-2}
\end{align}
Afterwards, node representations $h_v$ can be obtained by summing up all the incoming edge representations, as follows
\begin{align}
m_v &= \sum_{w\in N(v)} h_{vw}^T,\label{eq-3}\\
h_v &= \tau(W_o \text{ cat}(\mathbf{x}_v,m_v)). \label{eq-4}
\end{align}
Then, in the readout phase, a representation $h_G$ of the graph is obtained by summing or averaging the node representations. Finally, a feed-forward neural network $f$ is used to generate property predictions $\hat{y} = f(h_G)$. 

\subsubsection{Quotient Graphs}

Let $G$ be a graph and $\mathcal{P}=\{V_1, V_2, \ldots, V_k\}$ be a partition set of $V(G)$, i.e., for all $v\in V(G)$, there exists a unique $i \in \{1,2,\ldots,k\}$ such that $v \in V_i$. Then, the equivalence relation specified by the partition set $\mathcal{P}$ is defined as follows:
$$v\sim w \Longleftrightarrow \exists i \in \{1,2,\ldots,k\} \text{ s.t. } v,w \in V_i,$$
where $v,w\in V(G)$, and $v\sim w$ means that $v$ and $w$ are equivalent.

\begin{definition}\cite{Hahn}\label{def-qg}
Let $G$ be a graph and $\mathcal{P}=\{V_1, V_2, \ldots, V_k\}$ be a partition set of $V(G)$. The quotient graph $G/\mathcal{P}$ under the equivalence relation specified by $\mathcal{P}$ is a graph with the node set $\{V_1, V_2, \ldots, V_k\}$, such that for all distinct $i,j\in \{1,2,\ldots,k\}$ we have
$$V_iV_j\in E(G/\mathcal{P})\Longleftrightarrow \exists (v,w) \in (V_i,V_j) \text{ s.t. } vw\in E(G).$$
Each node $V_i$ of $G/\mathcal{P}$ is called a \textit{contracted node} from the set of nodes $V_i$ of $G$. Also, each edge $V_iV_j$ of $G/\mathcal{P}$ is called a \textit{contracted edge} from the set of edges $\{vw\in E(G): (v,w) \in (V_i,V_j)\}$.
\end{definition}
Figure \ref{fig-qg} gives an illustrative example for Definition \ref{def-qg}.

\subsection{The FunQG Framework} \label{FunQG}
The physical and chemical properties of compounds that contain a particular functional group tend to be similar~\cite{Headley}. Therefore by most theoretical studies, FGs are considered as keys to hierarchically classifying molecules~\cite{Feldman}. However, most cheminformatics based GNNs have paid no attention to FGs.

Let $\mathcal{M}$ be a molecule and $G_\mathcal{M}$ be its corresponding molecular graph. In this paper, a molecular graph coarsening framework, named \textbf{FunQG}, is proposed. This framework firstly constructs a partition set $\mathcal{P}$ of the node set $V(G_\mathcal{M})$ based on \textbf{fun}ctional groups of $\mathcal{M}$, and then considers the \textbf{q}uotient \textbf{g}raph of $G_\mathcal{M}$ under the equivalence relation specified by $\mathcal{P}$ as the molecular quotient graph of $G$. To avoid loss of information, the FunQG aggregates node and edge features of the original graph in their corresponding new nodes and edges, respectively, of the quotient graph. 

There are slightly different definitions for FGs by medicinal chemists. Also, extracting FGs from molecules may have high computational costs. Here, we utilize a relatively simple algorithm presented by Ertl~\cite{Ertl} to identify all FGs in organic molecules. This algorithm enables us to analyze FGs in large chemical databases very quickly and with low computational costs, which was not possible using other approaches. In this work, we utilize an open-source Python version of this algorithm written by Hall~\cite{Hall} to extract FGs from molecular datasets. More details about FGs and Ertl\textquotesingle s algorithm are deferred to the Supplementary Section 4.

\subsubsection{FunQG and Molecular Quotient Graphs} 
Let $\mathcal{M}$ be a molecule and $G_{\mathcal{M}}$ be its molecular graph. Suppose that there exist $m$ FGs in ${\mathcal{M}}$, and $\mathcal{FG}=\{S_1,S_2,\ldots,S_m\}$ is the set of all subgraphs of $G_{\mathcal{M}}$ induced by FGs of ${\mathcal{M}}$, i. e., for every FG $\mathcal{G}$ of ${\mathcal{M}}$ with corresponding atom set $A_\mathcal{G}$, there exists a unique $i\in \{1,2,\ldots,m\}$ such that the molecular subgraph induced by $A_\mathcal{G}$ is $S_i$. Then, the steps for applying the FunQG framework to $\mathcal{M}$ are as follows:

\begin{enumerate}
\item
Find $E_{cut}\subseteq E(G_{\mathcal{M}})$ such that $vw\in E_{cut}$ if and only if there exists an $i\in \{1,2,\ldots,m\}$ such that either $v$ or $w$ is a node in $S_i$. 

\item
Let $G_{cut}$ be the subgraph of $G_{\mathcal{M}}$ obtained from deletion of $E_{cut}$ from $E(G_{\mathcal{M}})$. Let $\{G_1, G_2, \ldots, G_c\}$ be the set of all connected components of $G_{cut}$. Consider $\{V(G_1), V(G_2), \ldots, V(G_c)\}$ as the partition set $\mathcal{P}$ of $V(G_\mathcal{M})$.

\item
Let $f$ be an aggregation function, e.g., sum, mean, etc. Find the quotient graph $G_{\mathcal{M}}/\mathcal{P}$ of the graph $G_{\mathcal{M}}$ under the equivalence relation specified by the partition set $\mathcal{P}$ of $V(G_{\mathcal{M}})$, such that the node features of each contracted node $V(G_i)$ in $G_{\mathcal{M}}/\mathcal{P}$ is equal to the output of $f$ on the set of all node features of $V(G_i)$ in $G_{\mathcal{M}}$. Also, the edge features of each contracted edge $V(G_i)V(G_j)\in E(G_{\mathcal{M}}/\mathcal{P})$ is equal to the output of $f$ on the set of all edge features of $\{vv'\in E_{cut} | (v,v') \in (V(G_i),V(G_j))\}$. The graph $G_{\mathcal{M}}/\mathcal{P}$ with the above properties is named the molecular quotient graph of $\mathcal{M}$.
\end{enumerate}

\subsubsection{Details About The FunQG}
The following facts are true about the FunQG framework:

\begin{enumerate}

\item
By the definition of FGs in the Ertl\textquotesingle s algorithm~\cite{Ertl}, two different FGs in a molecule have no atoms in common. Therefore, in the first step of the FunQG, if $v$ is a node in $S_i$, then there exists no $j\in \{1,2,\ldots,m\}\setminus \{i\}$ such that $v$ is a node in $S_j$.

\item
By the definition of FGs in the Ertl\textquotesingle s algorithm, for all $i\in \{1,2,\ldots,m\}$, $S_i$ is a connected subgraph of $G_{\mathcal{M}}$.

\item
By items $1$ and $2$, $\mathcal{FG}\subseteq \{G_1, G_2, \ldots, G_c\}$, and so $\mathcal{P_{FG}} = \{V(S_1),V(S_2),\ldots,V(S_m)\}\subseteq \mathcal{P}$. So, we have $\mathcal{P} = \mathcal{P_{FG}}\cup \mathcal{P_{\text{non-}FG}}$, where $\mathcal{P_{\text{non-}FG}} = \mathcal{P} \setminus \mathcal{P_{FG}}$. Let $V(\mathcal{FG}) = \cup_{i\in\{1,2,\ldots,m\}}V(S_i)$ and $V(\mathcal{\text{non-}FG})=V(G_{\mathcal{M}})\setminus V(\mathcal{FG})$. By the definition of FGs in the Ertl\textquotesingle s algorithm, each node in $V(\mathcal{\text{non-}FG})$ corresponds to a carbon atom.

\item
By item 3, since $\mathcal{P} = \mathcal{P_{FG}}\cup \mathcal{P_{\text{non-}FG}}$, the third step of the FunQG can be decomposed into two consecutive steps as follows:
\begin{enumerate}
\item
\textit{Quotient by non-FGs:} An aggregation function $f_1$ is considered and the third step of the FunQG is executed for $f_1$ and the partition set $\mathcal{P}'=\mathcal{P_{\text{non-}FG}}\cup \{\{v\} | v\in V(\mathcal{FG})\}$, to find the quotient graph $G_{\mathcal{M}}/\mathcal{P}'$ of $G_{\mathcal{M}}$.
\item
\textit{Quotient by FGs:} An aggregation function $f_2$ is considered and the third step of the FunQG is executed for $f_2$ and the partition set $\mathcal{P}'' = \mathcal{P_{FG}}\cup \{\{v\} | v\in V(G_{\mathcal{M}}/\mathcal{P}')\setminus V(\mathcal{FG})\}$ to find the quotient graph $\left(G_{\mathcal{M}}/\mathcal{P}'\right)/\mathcal{P}''$ of $G_{\mathcal{M}}/\mathcal{P}'$, where by $V(\mathcal{FG})$ here we mean the nodes of $G_{\mathcal{M}}/\mathcal{P}'$ corresponding to the nodes in $V(\mathcal{FG})$. Note that the quotient graph $\left(G_{\mathcal{M}}/\mathcal{P}'\right)/\mathcal{P}''$ is equal to the quotient graph $G_{\mathcal{M}}/\mathcal{P}$ of $G_{\mathcal{M}}$.
\end{enumerate}

\item
By item 4, the aggregation function $f_1$ may be different from the aggregation function $f_2$. In this paper, we utilize the mean aggregation function as $f_1$ (because by item 3, all nodes in $V(\mathcal{\text{non-}FG})$ correspond to carbon atoms) and the sum aggregation function as $f_2$. 

\item
By our experiments on several molecular property prediction benchmarks from the MoleculeNet~\cite{Wu2}, on average, the resulting molecular quotient graphs are much smaller than the molecular graphs. For a molecular dataset, let the \textit{abstraction ratio} be the total number of nodes in molecular quotient graphs divided by the total number of nodes in molecular graphs. It can be seen that for each dataset of our experiment, the abstraction ratio is at most $0.38$. In Section \ref{sec-results}, we see that applying the FunQG framework to molecular property prediction benchmarks not only reduces the required memory and computational costs of GNNs but also achieves significant performance. More details about the size of the molecular graphs, the size of the molecular quotient graphs, and FGs of the datasets can be found in the Supplementary Section 4. 

\end{enumerate}

\begin{figure*}[!t]
\begin{center}
\hspace*{-1em}
\includegraphics[scale=0.53]{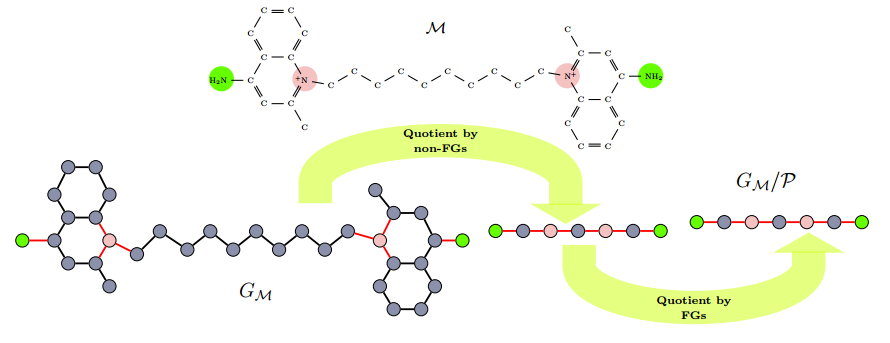}
\vspace*{0.3em}
\caption{Application of the FunQG to molecule $\mathcal{M}$ with the SMILES string Cc1cc(N)c2ccccc2[n+]1CCCCCCCCCC[n+]1c(C)cc(N)c2ccccc21 from the Tox21 dataset results in the molecular quotient graph $G_{\mathcal{M}}/ \mathcal{P}$. Here, there exists no difference between the graphs before and after the "Quotient by FGs" operator because the FGs of $\mathcal{M}$ are all single atoms.}\label{fig-qg-1}


\hspace*{-1em}
\includegraphics[scale=0.47]{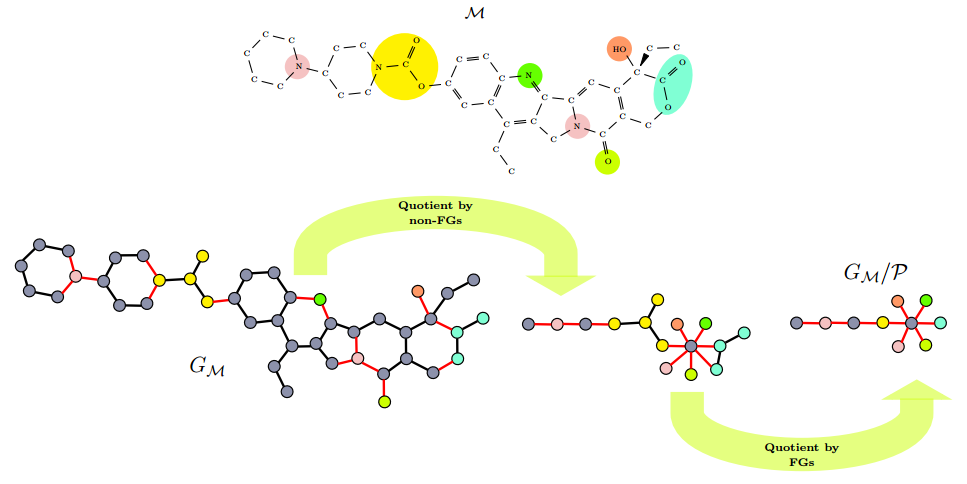}
\vspace*{0.3em}
\end{center}
\caption{Application of the FunQG to molecule $\mathcal{M}$ with the SMILES string CCc1c2c(nc3ccc(OC(=O)N4CCC(N5CCCCC5)CC4)cc13)-c1cc3c(c(=O)n1C2)COC(=O)[C@]3(O)CC from the Tox21 dataset results in the molecular quotient graph $G_{\mathcal{M}}/ \mathcal{P}$.}\label{fig-qg-3}
\end{figure*}

Figures \ref{fig-qg-2}, \ref{fig-qg-1}, and \ref{fig-qg-3} give illustrative examples for the FunQG framework. In these three figures, edges in $E_{cut}$ are drawn in red. Figure \ref{fig-qg-1} shows no difference between the graphs before and after the "Quotient by FGs" operator because the FGs of the related molecule are all single atoms.

\begin{figure}[!t]
\begin{center}
{\scalebox{0.7}{
\begin{tikzpicture}[x=0.75pt,y=0.75pt,yscale=-1,xscale=1]
\draw [line width=1.5]  [dash pattern={on 1.69pt off 2.76pt}]  (0,40) -- (40.14,50.95) ;
\draw [shift={(44,52)}, rotate = 195.26] [fill={rgb, 255:red, 0; green, 0; blue, 0 }  ][line width=0.08]  [draw opacity=0] (15.72,-7.55) -- (0,0) -- (15.72,7.55) -- (10.44,0) -- cycle    ;
\draw [line width=1.5]  [dash pattern={on 1.69pt off 2.76pt}]  (0,73) -- (38,73.9) ;
\draw [shift={(42,74)}, rotate = 181.36] [fill={rgb, 255:red, 0; green, 0; blue, 0 }  ][line width=0.08]  [draw opacity=0] (15.72,-7.55) -- (0,0) -- (15.72,7.55) -- (10.44,0) -- cycle    ;
\draw [line width=1.5]    (73,76) -- (112.07,112.28) ;
\draw [shift={(115,115)}, rotate = 222.88] [fill={rgb, 255:red, 0; green, 0; blue, 0 }  ][line width=0.08]  [draw opacity=0] (15.72,-7.55) -- (0,0) -- (15.72,7.55) -- (10.44,0) -- cycle    ;
\draw [line width=1.5]    (78,124) -- (109.01,125.77) ;
\draw [shift={(113,126)}, rotate = 183.27] [fill={rgb, 255:red, 0; green, 0; blue, 0 }  ][line width=0.08]  [draw opacity=0] (15.72,-7.55) -- (0,0) -- (15.72,7.55) -- (10.44,0) -- cycle    ;
\draw [line width=1.5]  [dash pattern={on 1.69pt off 2.76pt}]  (0,104) -- (39.07,111.27) ;
\draw [shift={(43,112)}, rotate = 190.54] [fill={rgb, 255:red, 0; green, 0; blue, 0 }  ][line width=0.08]  [draw opacity=0] (15.72,-7.55) -- (0,0) -- (15.72,7.55) -- (10.44,0) -- cycle    ;
\draw [line width=1.5]  [dash pattern={on 1.69pt off 2.76pt}]  (1,141) -- (38.09,132.86) ;
\draw [shift={(42,132)}, rotate = 167.62] [fill={rgb, 255:red, 0; green, 0; blue, 0 }  ][line width=0.08]  [draw opacity=0] (15.72,-7.55) -- (0,0) -- (15.72,7.55) -- (10.44,0) -- cycle    ;
\draw  [color={rgb, 255:red, 0; green, 34; blue, 73 }  ,draw opacity=1 ][fill=brightgreen  ,fill opacity=1 ] (43.5,192.5) .. controls (43.5,182.01) and (52.23,173.5) .. (63,173.5) .. controls (73.77,173.5) and (82.5,182.01) .. (82.5,192.5) .. controls (82.5,202.99) and (73.77,211.5) .. (63,211.5) .. controls (52.23,211.5) and (43.5,202.99) .. (43.5,192.5) -- cycle ;
\draw [line width=1.5]  [dash pattern={on 1.69pt off 2.76pt}]  (1.5,170.5) -- (42.66,182.39) ;
\draw [shift={(46.5,183.5)}, rotate = 196.11] [fill={rgb, 255:red, 0; green, 0; blue, 0 }  ][line width=0.08]  [draw opacity=0] (15.72,-7.55) -- (0,0) -- (15.72,7.55) -- (10.44,0) -- cycle    ;
\draw [line width=1.5]  [dash pattern={on 1.69pt off 2.76pt}]  (1.28,204.5) -- (41.51,201.77) ;
\draw [shift={(45.5,201.5)}, rotate = 176.12] [fill={rgb, 255:red, 0; green, 0; blue, 0 }  ][line width=0.08]  [draw opacity=0] (15.72,-7.55) -- (0,0) -- (15.72,7.55) -- (10.44,0) -- cycle    ;
\draw [line width=1.5]    (82.5,192.5) -- (108.5,193.37) ;
\draw [shift={(112.5,193.5)}, rotate = 181.91] [fill={rgb, 255:red, 0; green, 0; blue, 0 }  ][line width=0.08]  [draw opacity=0] (15.72,-7.55) -- (0,0) -- (15.72,7.55) -- (10.44,0) -- cycle    ;
\draw [line width=1.5]    (74.5,244.5) -- (113.6,207.26) ;
\draw [shift={(116.5,204.5)}, rotate = 136.4] [fill={rgb, 255:red, 0; green, 0; blue, 0 }  ][line width=0.08]  [draw opacity=0] (15.72,-7.55) -- (0,0) -- (15.72,7.55) -- (10.44,0) -- cycle    ;
\draw [line width=1.5]  [dash pattern={on 1.69pt off 2.76pt}]  (1.5,246.5) -- (40.51,249.22) ;
\draw [shift={(44.5,249.5)}, rotate = 183.99] [fill={rgb, 255:red, 0; green, 0; blue, 0 }  ][line width=0.08]  [draw opacity=0] (15.72,-7.55) -- (0,0) -- (15.72,7.55) -- (10.44,0) -- cycle    ;
\draw [line width=1.5]  [dash pattern={on 1.69pt off 2.76pt}]  (2.5,277.5) -- (40.59,269.34) ;
\draw [shift={(44.5,268.5)}, rotate = 167.91] [fill={rgb, 255:red, 0; green, 0; blue, 0 }  ][line width=0.08]  [draw opacity=0] (15.72,-7.55) -- (0,0) -- (15.72,7.55) -- (10.44,0) -- cycle    ;
\draw  [color={rgb, 255:red, 0; green, 34; blue, 73 }  ,draw opacity=1 ][fill=brightgreen  ,fill opacity=1 ][line width=0.75]  (39,63) .. controls (39,52.51) and (47.73,44) .. (58.5,44) .. controls (69.27,44) and (78,52.51) .. (78,63) .. controls (78,73.49) and (69.27,82) .. (58.5,82) .. controls (47.73,82) and (39,73.49) .. (39,63) -- cycle ;
\draw  [color={rgb, 255:red, 0; green, 34; blue, 73 }  ,draw opacity=1 ][fill=brightgreen  ,fill opacity=1 ] (39,124) .. controls (39,113.51) and (47.73,105) .. (58.5,105) .. controls (69.27,105) and (78,113.51) .. (78,124) .. controls (78,134.49) and (69.27,143) .. (58.5,143) .. controls (47.73,143) and (39,134.49) .. (39,124) -- cycle ;
\draw  [color={rgb, 255:red, 0; green, 34; blue, 73 }  ,draw opacity=1 ][fill=brightgreen  ,fill opacity=1 ] (41.5,258.5) .. controls (41.5,248.01) and (50.23,239.5) .. (61,239.5) .. controls (71.77,239.5) and (80.5,248.01) .. (80.5,258.5) .. controls (80.5,268.99) and (71.77,277.5) .. (61,277.5) .. controls (50.23,277.5) and (41.5,268.99) .. (41.5,258.5) -- cycle ;
\draw  [color={rgb, 255:red, 0; green, 34; blue, 73 }  ,draw opacity=1 ][fill=brightgreen  ,fill opacity=1 ] (112,125) .. controls (112,114.51) and (120.73,106) .. (131.5,106) .. controls (142.27,106) and (151,114.51) .. (151,125) .. controls (151,135.49) and (142.27,144) .. (131.5,144) .. controls (120.73,144) and (112,135.49) .. (112,125) -- cycle ;
\draw  [color={rgb, 255:red, 0; green, 34; blue, 73 }  ,draw opacity=1 ][fill=brightgreen  ,fill opacity=1 ] (112.5,193.5) .. controls (112.5,183.01) and (121.23,174.5) .. (132,174.5) .. controls (142.77,174.5) and (151.5,183.01) .. (151.5,193.5) .. controls (151.5,203.99) and (142.77,212.5) .. (132,212.5) .. controls (121.23,212.5) and (112.5,203.99) .. (112.5,193.5) -- cycle ;
\draw [color={rgb, 255:red, 0; green, 0; blue, 0 }  ,draw opacity=1 ][line width=1.5]    (148.5,182.5) -- (184.7,170.74) ;
\draw [shift={(188.5,169.5)}, rotate = 162] [fill={rgb, 255:red, 0; green, 0; blue, 0 }  ,fill opacity=1 ][line width=0.08]  [draw opacity=0] (15.72,-7.55) -- (0,0) -- (15.72,7.55) -- (10.44,0) -- cycle    ;
\draw [color={rgb, 255:red, 0; green, 0; blue, 0 }  ,draw opacity=1 ][line width=1.5]    (146,138) -- (184.15,148.9) ;
\draw [shift={(188,150)}, rotate = 195.95] [fill={rgb, 255:red, 0; green, 0; blue, 0 }  ,fill opacity=1 ][line width=0.08]  [draw opacity=0] (15.72,-7.55) -- (0,0) -- (15.72,7.55) -- (10.44,0) -- cycle    ;
\draw  [color={rgb, 255:red, 0; green, 34; blue, 73 }  ,draw opacity=1 ][fill=brightgreen  ,fill opacity=1 ] (184.5,160) .. controls (184.5,149.51) and (193.23,141) .. (204,141) .. controls (214.77,141) and (223.5,149.51) .. (223.5,160) .. controls (223.5,170.49) and (214.77,179) .. (204,179) .. controls (193.23,179) and (184.5,170.49) .. (184.5,160) -- cycle ;
\draw [color=atomictangerine  ,draw opacity=1 ][line width=1.5]    (0,41) .. controls (164.05,68.49) and (71.81,160.21) .. (269.5,149.17) ;
\draw [shift={(272.5,149)}, rotate = 176.56] [fill=atomictangerine  ,fill opacity=1 ][line width=0.08]  [draw opacity=0] (11.61,-5.58) -- (0,0) -- (11.61,5.58) -- cycle    ;
\draw [color=atomictangerine  ,draw opacity=1 ][line width=1.5]    (0,81) .. controls (116.8,48.29) and (49.68,161.85) .. (267.69,152.16) ;
\draw [shift={(271,152)}, rotate = 177.16] [fill=atomictangerine  ,fill opacity=1 ][line width=0.08]  [draw opacity=0] (11.61,-5.58) -- (0,0) -- (11.61,5.58) -- cycle    ;
\draw [color=atomictangerine  ,draw opacity=1 ][line width=1.5]    (0,100.5) .. controls (106.47,99.51) and (92.64,162.37) .. (267.3,155.29) ;
\draw [shift={(269.95,155.18)}, rotate = 177.48] [fill=atomictangerine  ,fill opacity=1 ][line width=0.08]  [draw opacity=0] (11.61,-5.58) -- (0,0) -- (11.61,5.58) -- cycle    ;
\draw [color=atomictangerine  ,draw opacity=1 ][line width=1.5]    (0,277.13) .. controls (155.1,262.7) and (73.21,157.19) .. (270.5,170.29) ;
\draw [shift={(273.5,170.5)}, rotate = 184.09] [fill=atomictangerine  ,fill opacity=1 ][line width=0.08]  [draw opacity=0] (11.61,-5.58) -- (0,0) -- (11.61,5.58) -- cycle    ;
\draw [color=atomictangerine  ,draw opacity=1 ][line width=1.5]    (0,238) .. controls (132.68,275.44) and (47.24,154.84) .. (268.64,167.79) ;
\draw [shift={(272,168)}, rotate = 183.64] [fill=atomictangerine  ,fill opacity=1 ][line width=0.08]  [draw opacity=0] (11.61,-5.58) -- (0,0) -- (11.61,5.58) -- cycle    ;
\draw [color=atomictangerine  ,draw opacity=1 ][line width=1.5]    (0,204) .. controls (99.51,228.38) and (88.49,157.84) .. (267.78,164.4) ;
\draw [shift={(270.5,164.5)}, rotate = 182.32] [fill=atomictangerine  ,fill opacity=1 ][line width=0.08]  [draw opacity=0] (11.61,-5.58) -- (0,0) -- (11.61,5.58) -- cycle    ;
\draw [color=atomictangerine  ,draw opacity=1 ][line width=1.5]    (0,174) .. controls (120.4,208.83) and (87.22,158.64) .. (267.22,161.63) ;
\draw [shift={(269.95,161.68)}, rotate = 181.11] [fill=atomictangerine  ,fill opacity=1 ][line width=0.08]  [draw opacity=0] (11.61,-5.58) -- (0,0) -- (11.61,5.58) -- cycle    ;
\draw [color=atomictangerine  ,draw opacity=1 ][line width=1.5]    (0,136) .. controls (127.74,99.81) and (57.52,160.07) .. (266.77,158.7) ;
\draw [shift={(269.95,158.68)}, rotate = 179.46] [fill=atomictangerine  ,fill opacity=1 ][line width=0.08]  [draw opacity=0] (11.61,-5.58) -- (0,0) -- (11.61,5.58) -- cycle    ;
\draw  [color={rgb, 255:red, 0; green, 0; blue, 0 }  ,draw opacity=0 ][fill=aquamarine  ,fill opacity=1 ] (251.25,111.93) -- (251.25,119.72) -- (262,119.72) -- (240.5,137.5) -- (219,119.72) -- (229.75,119.72) -- (229.75,111.93) -- cycle ;\draw  [color={rgb, 255:red, 0; green, 0; blue, 0 }  ,draw opacity=0 ][fill=aquamarine  ,fill opacity=1 ] (251.25,103.41) -- (251.25,105.12) -- (229.75,105.12) -- (229.75,103.41) -- cycle ;\draw  [color={rgb, 255:red, 0; green, 0; blue, 0 }  ,draw opacity=0 ][fill=aquamarine  ,fill opacity=1 ] (251.25,106.82) -- (251.25,110.23) -- (229.75,110.23) -- (229.75,106.82) -- cycle ;
\draw [color={rgb, 255:red, 0; green, 0; blue, 0 }  ,draw opacity=1 ][line width=1.5]    (223.5,160) -- (265.95,159.25) ;
\draw [shift={(269.95,159.18)}, rotate = 178.98] [fill={rgb, 255:red, 0; green, 0; blue, 0 }  ,fill opacity=1 ][line width=0.08]  [draw opacity=0] (15.72,-7.55) -- (0,0) -- (15.72,7.55) -- (10.44,0) -- cycle    ;
\draw  [color={rgb, 255:red, 0; green, 34; blue, 73 }  ,draw opacity=1 ][fill=brightgreen  ,fill opacity=1 ] (269.95,158.68) .. controls (269.95,148.18) and (278.68,139.68) .. (289.45,139.68) .. controls (300.22,139.68) and (308.95,148.18) .. (308.95,158.68) .. controls (308.95,169.17) and (300.22,177.68) .. (289.45,177.68) .. controls (278.68,177.68) and (269.95,169.17) .. (269.95,158.68) -- cycle ;
\draw (157,60) node [anchor=north west][inner sep=0.75pt]   [align=left] {\begin{minipage}[lt]{114.44pt}\setlength\topsep{0pt}
\begin{center}
\textbf{{\Large Over-squashing}}\\\textbf{{\Large bottleneck}}
\end{center}
\end{minipage}};
\end{tikzpicture}
}}
\caption{Over-squashing bottleneck of graph neural networks~\cite{Alon}} \label{fig-over-squash}
\end{center}
\end{figure}
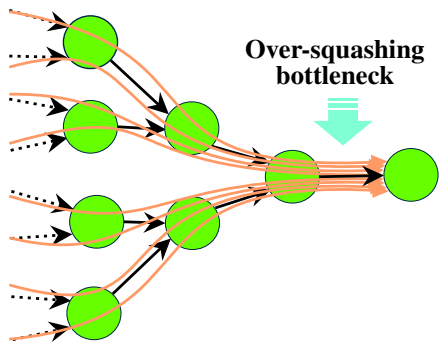

Alon~et~al.~\cite{Alon} show that over-squashing of information from exponentially many neighbors into fixed-size vectors is a performance bottleneck in tasks that require long-range interaction and so more number of GNN layers, such as molecular property prediction tasks, which may depend on the combination of atoms that reside in the molecule’s opposite sides (Figure \ref{fig-over-squash}). The molecular graph $G_\mathcal{M}$ in Figure \ref{fig-qg-1} is an example of a graph that may suffer the over-squashing issue during a message passing process. Figure \ref{fig-qg-1} is a good example to show how using the FunQG for graph coarsening can alleviate the over-squashing issue of GNNs.

\section{Results}\label{sec-results}
To evaluate the efficiency of the FunQG, we apply it to several molecular property prediction benchmarks, and then compare the performances of a GNN model architecture on the obtained datasets with the performances of several state-of-the-art baselines on the original datasets. To demonstrate the effectiveness of our proposed framework, and to maintain the simplicity of the method to reduce computational costs, we employ MPNN and DMPNN architectures as two popular simple baseline GNNs for predicting molecular properties, which need no attention layers or pre-training. We denote these combinations by FunQG-MPNN and FunQG-DMPNN, respectively.

\subsection{Datasets and Splitting Method}
We conduct experiments on several popular benchmark datasets from MoleculeNet, including both classification and regression tasks~\cite{Wu2}. More details about the datasets can be found in the Supplementary Section 1.

Although \textit{scaffold splitting} is more challenging than \textit{random splitting}, using it for molecular property prediction tasks can better evaluate out-of-distribution generalisation abilities of the models~\cite{Wu2}. Following the previous works, e.g.,~\cite{Li2, Rong}, we perform three independent executions on three randomly seeded scaffold splittings with the ratio of 80:10:10 for the training/validation/test sets, and then report the means and standard deviations of the evaluation metrics. More details about this splitting method can be found in the Supplementary Section 2.

\subsection{Baselines}
We compare the performance of the proposed method with various baselines. The considered baseline models and their performances are taken from recent works on molecular representation learning, such as~\cite{Li2},~\cite{Rong}, and~\cite{Ma}. Among these baselines, ECFP~\cite{Rogers} and TF\textunderscore Robust~\cite{Ramsundar} are non-graph models taking the molecular fingerprints as inputs. But, GraphConv~\cite{Kipf}, Weave~\cite{Kearnes}, SchNet~\cite{Schutt}, MPNN~\cite{Gilmer}, DMPNN~\cite{Yang}, MGCN~\cite{Lu}, AttentiveFP~\cite{Xiong} and TrimNet~\cite{Li3} are GNN-based models. More specifically, GraphConv, Weave and SchNet are graph convolutional models, while MPNN, DMPNN, MGCN, AttentiveFP and TrimNet are GNN models considering the edge features during message passing. Furthermore, AttentiveFP and TrimNet are variants of the graph attention networks. 

\subsection{Evaluation Metrics}
Following the evaluation metrics used by most recent baseline models, in this paper, all classification models are evaluated by \textit{ROC-AUC} (area under the receiver operating characteristic curve), and regression models are evaluated by \textit{RMSE} (root-mean-square error).

\subsection{Experiments Settings}

\subsubsection{GNN Architecture}\label{sec-gnn}
In our experiments, firstly, we apply the FunQG to a molecular dataset. Then, we utilize MPNN and DMPNN as two GNN architectures on the obtained dataset to evaluate the performance of the proposed graph coarsening framework. For MPNN and DMPNN, our implementations are the same as those of Yang~et al.~\cite{Yang}. Also, in both models, we use ReLU as activation function, and in the readout phase, we use average pooling to obtain the graphs representations. Then, a feed-forward neural network with two hidden layers and the activation function ReLU is utilized to generate property predictions.

\newcommand{\ra}[1]{\renewcommand{\arraystretch}{#1}}
\begin{table*}[t]
\centering
\caption{Performance comparison on $9$ benchmarks}\label{tab-result}
\ra{6}
\fontsize{25}{7}\selectfont
\resizebox{\textwidth}{!}{
\begin{tabular}{>{\kern-\tabcolsep}lcccccccccc<{\kern-\tabcolsep}}
\toprule
& \multicolumn{6}{c}{Classification (higher is better)} & \phantom{a} & \multicolumn{3}{c}{Regression (lower is better)}\\
\cmidrule{2-7} \cmidrule{9-11} \\
\addlinespace[-1em]
{} & Tox21 & ToxCast  & ClinTox & SIDER & BBBP & BACE && FreeSolv & ESOL & Lipophilicity \\
{} & $7831$ & $8576$ & $1478$ & $1427$ & $2039$ & $1513$ && $642$ & $1128$ & $4200$ \\
\midrule
ECFP~\cite{Rogers} & $0.760(0.009)$ & $0.615(0.017)$ & $0.673(0.031)$ & $0.630(0.019)$ & $0.783(0.050)$ & $0.861(0.024)$ && $5.275(0.751)$ & $2.359(0.454)$ & $1.188(0.061)$\\
TF\textunderscore Robust~\cite{Ramsundar} & $0.698(0.012)$ & $0.585(0.031)$ & $0.765(0.085)$ & $0.607(0.033)$ & $0.860(0.087)$ & $0.824(0.022)$ && $4.122(0.085)$ & $1.722(0.038)$ & $0.909(0.060)$\\
GraphConv~\cite{Kipf} & $0.772(0.041)$ & $0.650(0.025)$ & $0.845(0.051)$ & $0.593(0.035)$ & $0.877(0.036)$ & $0.854(0.011)$ && $2.900(0.135)$ & $1.068(0.050)$ & $0.712(0.049)$\\
Weave~\cite{Kearnes} & $0.741(0.044)$ & $0.678(0.024)$ & $0.823(0.023)$ & $0.543(0.034)$ & $0.837(0.065)$ & $0.791(0.008)$ && $2.398(0.250)$ & $1.158(0.055)$ & $0.813(0.042)$\\
SchNet~\cite{Schutt} & $0.767(0.025)$ & $0.679(0.021)$ & $0.717(0.042)$ & $0.545(0.038)$ & $0.847(0.024)$ & $0.750(0.033)$ && $3.215(0.755)$ & $1.045(0.064)$ & $0.909(0.098)$\\
MGCN~\cite{Lu} & $0.707(0.016)$ & $0.663(0.009)$ & $0.634(0.042)$ & $0.552(0.018)$ & $0.850(0.064)$ & $0.734(0.030)$ && $3.349(0.097)$ & $1.266(0.147)$ & $1.113(0.041)$\\
AttentiveFP~\cite{Xiong} & $0.807(0.020)$ & $0.579(0.001)$ & $\mathbf{0.933(0.020)}$ & $0.605(0.060)$ & $0.908(0.050)$ & $\mathbf{0.863(0.015)}$ && $2.030(0.420)$ & $0.853(0.060)$ & $0.650(0.030)$\\
TrimNet~\cite{Li3} & $0.812(0.019)$ & $0.652(0.032)$ & $0.906(0.017)$ & $0.606(0.006)$ & $0.892(0.025)$ & $0.843(0.025)$ && $2.529(0.111)$ & $1.282(0.029)$ & $0.702(0.008)$\\
\cellcolor{cyan!30}
MPNN~\cite{Gilmer} & $0.808(0.024)$ & $0.691(0.013)$ & $0.879(0.054)^*$ & $0.595(0.030)$ & $0.913(0.041)^*$ & $0.815(0.044)$ && $2.185(0.952)$ & $1.167(0.430)$ & $0.672(0.051)$\\
\cellcolor{lightgray!60}
DMPNN~\cite{Yang} & $0.826(0.023)$ & $0.718(0.011)$ & $0.897(0.040)^{**}$ & $0.632(0.023)$ & $\mathbf{0.919(0.030)}^{**}$ & $0.852(0.053)$ && $2.177(0.914)$ & $0.980(0.258)$ & $0.653(0.046)$\\
\addlinespace[-1.3pt]
\bottomrule
\cellcolor{cyan!30}
FunQG-MPNN & $0.842(0.012)^*$ & $0.717(0.005)^*$ & $0.838(0.025)$ & $0.632(0.056)^*$ & $0.902(0.014)$ & $0.850(0.055)^*$ && $1.542(0.460)^*$ & $0.879(0.091)^*$ & $0.638(0.020)^*$ \\
\cellcolor{lightgray!60}
FunQG-DMPNN & $\mathbf{0.845(0.008)}^{**}$ & $\mathbf{0.721(0.009)}^{**}$ & $0.841(0.037)$ & $\mathbf{0.642(0.034)}^{**}$ & $0.914(0.010)$ & $0.862(0.047)^{**}$ && $\mathbf{1.501(0.376)}^{**}$ & $\mathbf{0.818(0.047)}^{**}$ & $\mathbf{0.622(0.028)}^{**}$ \\
\addlinespace[-1.3pt]
\bottomrule
\end{tabular}}
\begin{tablenotes}%
\item $1. $ 
We report the mean of three independent executions on three randomly seeded scaffold splittings with the ratio of 80:10:10 for the training/validation/test sets. The numbers in brackets are the standard deviations, and the best result in each dataset is written in bold. 
\item $2. $ 
The last two rows indicate the results of our methods. Blue (gray) shading cells are MPNN and FunQG-MPNN (DMPNN and FunQG-DMPNN). The best results in comparison between MPNN and FunQG-MPNN (DMPNN and FunQG-DMPNN) are shown with * (**). \vspace*{6pt}
\end{tablenotes}
\end{table*}

\subsubsection{Training and Hyperparameters Details}
We utilize the gradient descent and back-propagation algorithms by PyTorch~\cite{Paszke} and DGL~\cite{Wang} libraries to implement the models. Specifically, we use the \textit{Adam} optimizer to update the parameters in both training and hyperparameter tuning procedures. Also, we utilize three regularization techniques, including dropout, early stopping, and max-norm to prevent overfitting issues. More details about the training are deferred to the Supplementary Section 5.

\begin{figure}[t]
\centering
\pgfplotsset{%
    width=.4\textwidth,
    height=.6\textwidth
}
\begin{tikzpicture}
  \begin{axis}[
    xbar = .0cm,
    y axis line style = {opacity = 0},
    axis x line       = none,
    tickwidth         = 0pt,
    bar width		  = 3mm,
    tick label style={font=\footnotesize},
    legend style={at={(1.18,0.21)},anchor=north east, font=\scriptsize},
    label style={font=\footnotesize},
    ytick             = data,
    enlarge y limits  = 0.2,
    enlarge x limits  = 0.02,
    symbolic y coords = {Tox21,ToxCast,ClinTox,SIDER,BBBP,BACE,FreeSolv,ESOL,Lipophilicity},
    nodes near coords,
    nodes near coords style={color=black, font=\scriptsize},
    area legend,    
  ]
  \addplot[orange,fill=orange] coordinates {(124252,Tox21)(398777,ToxCast)(214262,ClinTox)(201247,SIDER)(175161,BBBP)(186511,BACE)(186291,FreeSolv)(210351,ESOL)(214561,Lipophilicity)}; 
  \addplot[blue-green,fill=blue-green] coordinates {(1390612,Tox21)(1254617,ToxCast)(11049402,ClinTox)(4351227,SIDER)(2282001,BBBP)(1069201,BACE)(10131001,FreeSolv)(2143501,ESOL)(3668101,Lipophilicity)}; 
  \legend{FunQG-DMPNN, DMPNN}
  \end{axis}
\end{tikzpicture}
\caption{A comparison between the total number of parameters of the best models \\ of the FunQG-DMPNN and the DMPNN}\label{fig-parameters}
\end{figure}
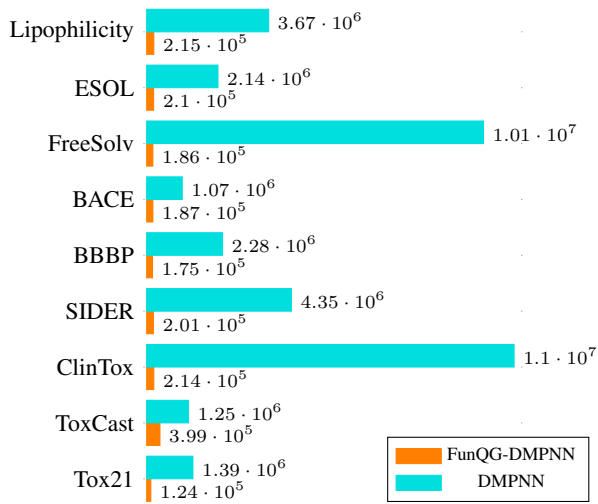

The resulting graphs of the FunQG are much smaller than the molecular graphs (Supplementary Section 4). Therefore, a GNN model architecture requires much less depth in working with the resulting graphs compared to working with the molecular graphs. Thus, using the FunQG reduces the computational costs. We utilize one \textit{Intel (R) Xeon (R) E5-2699 v4 @ 2.20GHz} CPU for training, testing, and hyperparameter tuning of the FunQG-MPNN and FunQG-DMPNN on each dataset in a relatively short time. For all tasks, Figure \ref{fig-parameters} compares the total number of parameters of the best models of the FunQG-DMPNN and the DMPNN. Information about the models of the DMPNN are taken from~\cite{Swanson}.

We perform \textit{Bayesian optimization} using the Ray Tune~\cite{Liaw} Python package to find the optimal hyperparameters for the FunQG-MPNN and FunQG-DMPNN. More details about the hyperparameters optimization process can be found in the Supplementary Section 5.

\subsection{Experimental Results}
Table \ref{tab-result} summarizes the performances of our proposed method alongside various baseline methods. This table indicates that the FunQG-DMPNN outperforms popular baselines on 6 of the 9 datasets, besides its low computational costs. Also, relatively small standard deviations indicate the robustness of our method. Furthermore by Table \ref{tab-result}, compared to the MPNN (DMPNN), the FunQG-MPNN (FunQG-DMPNN) performs better in 7 out of 9 studied benchmarks. 

\section{Conclusion and Future Work}
In this paper, we have proposed a novel molecular graph coarsening framework named FunQG for more efficient learning of molecular representations. Our method focuses on obtaining richer, more effective, and less challenging graphs for training graph neural networks from molecular graphs. To achieve this goal, we used a concept of graph theory called quotient graphs along with important molecular substructures called functional groups. Since selecting appropriate graph partitions is crucial to avoid information loss in constructing quotient graphs, we utilized FGs for this purpose. A direction for future research is to consider other domain knowledge or graph-theoretic information about the molecular graphs to partition their node sets. On the other hand, medicinal chemists have relatively different definitions of FGs. In this paper, we employed a relatively simple algorithm to identify FGs in large chemical databases very quickly and with low computational costs. Another future research direction is to use other definitions and algorithms to determine FGs. Also, we respectively utilized mean and sum functions to aggregate local information of molecular graphs into the nodes and edges of the quotient graphs to avoid data loss. One can use other combinations of aggregation functions in a similar manner. Through experiments, we showed that the resulting informative graphs are much smaller than the molecular graphs and thus are good candidates for making GNNs more efficient for finding molecular representations. Since the FunQG framework is model-independent, one can try different models on the resulting graphs to find an optimal model architecture. To demonstrate the efficiency of the FunQG, and to maintain the simplicity and cost-effectiveness of the method, we used the MPNN and DMPNN architectures as two simple state-of-the-art baselines for molecular property prediction, which need no attention layers or pre-training. According to our experiments, this method significantly outperforms previous baselines on various popular molecular property prediction benchmarks despite its simplicity and low computational costs. 

\hspace*{0em}
\setlength\fboxsep{0.5cm}
\fbox{\parbox{0.93\textwidth}{
\textbf{Key Points}
\begin{itemize}
\setlength{\itemindent}{-0.5em}
\item 
FunQG is a novel molecular graph coarsening framework for more efficient learning of molecular representations.
\item
FunQG incorporates domain knowledge and graph theory to obtain more efficient and less challenging candidates for training GNNs.
\item 
FunQG simplifies molecular graphs into small and informative ones to overcome some fundamental limitations of GNNs, rooted in working with relatively large-size graphs.
\item 
The experimental results on different property prediction benchmarks using the FunQG and two GNN architectures show that this method significantly outperforms previous baselines on various datasets despite its simplicity and low computational costs.
\item 
FunQG can be used as a simple, cost-effective, and robust method for solving the molecular representation learning problem.
\end{itemize}
}}

\section{Data and Code Availability}
The original molecular datasets are available on the website of MoleculeNet at \url{http://moleculenet.org}. The source code and data underlying this work are available in GitHub at \url{https://github.com/hhaji/funqg}.

\section{Competing interests}
The authors declare no competing financial interest.


\section{Funding}
Hossein Hajiabolhassan was supported by a grant from the National Natural Science Foundation of China (NSFC-U20A2068). Also, Zahra Taheri was supported by a grant from Basic Sciences Research Fund of Ministry of Science, Research, and Technology of Iran (BSRF-math-399-06).

\end{document}